\documentclass[10pt,twocolumn,letterpaper]{article}
\usepackage{authblk}

\renewcommand\thefootnote{\fnsymbol{footnote}}

\usepackage[pagenumbers]{cvpr} 

\definecolor{cvprblue}{rgb}{0.21,0.49,0.74}
\usepackage[pagebackref,breaklinks,colorlinks,allcolors=cvprblue]{hyperref}
\usepackage{multirow}
\usepackage{float}
\usepackage{xcolor}
\usepackage{stfloats}
\def\paperID{*****} 
\def\confName{CVPR}
\def\confYear{2027}

\title{SwiftVR: Real-Time One-Step Generative Video Restoration}

\author[1,2]{Jiaqi Yan\protect\footnotemark[1]\protect\footnotemark[2]}
\author[2]{Xiangyu Chen\protect\footnotemark[1]}
\author[2,3]{Xinlin Zhong}
\author[2]{Haibin Huang}
\author[2]{Chi Zhang}
\author[3]{\protect\\Jie Liu}
\author[1]{Jiantao Zhou\protect\footnotemark[3]}
\author[2]{Xuelong Li\protect\footnotemark[3]}

\affil[1]{State Key Laboratory of Internet of Things for Smart City, \protect\\
Department of Computer and Information Science, University of Macau}
\affil[2]{Institute of Artificial Intelligence (TeleAI), China Telecom}
\affil[3]{State Key Laboratory for Novel Software Technology, Nanjing University}

\begin{document}



\makeatletter
\twocolumn[{%
  \@maketitle
  \vspace{-2.5em}
  \begin{center}
    \includegraphics[width=\textwidth]{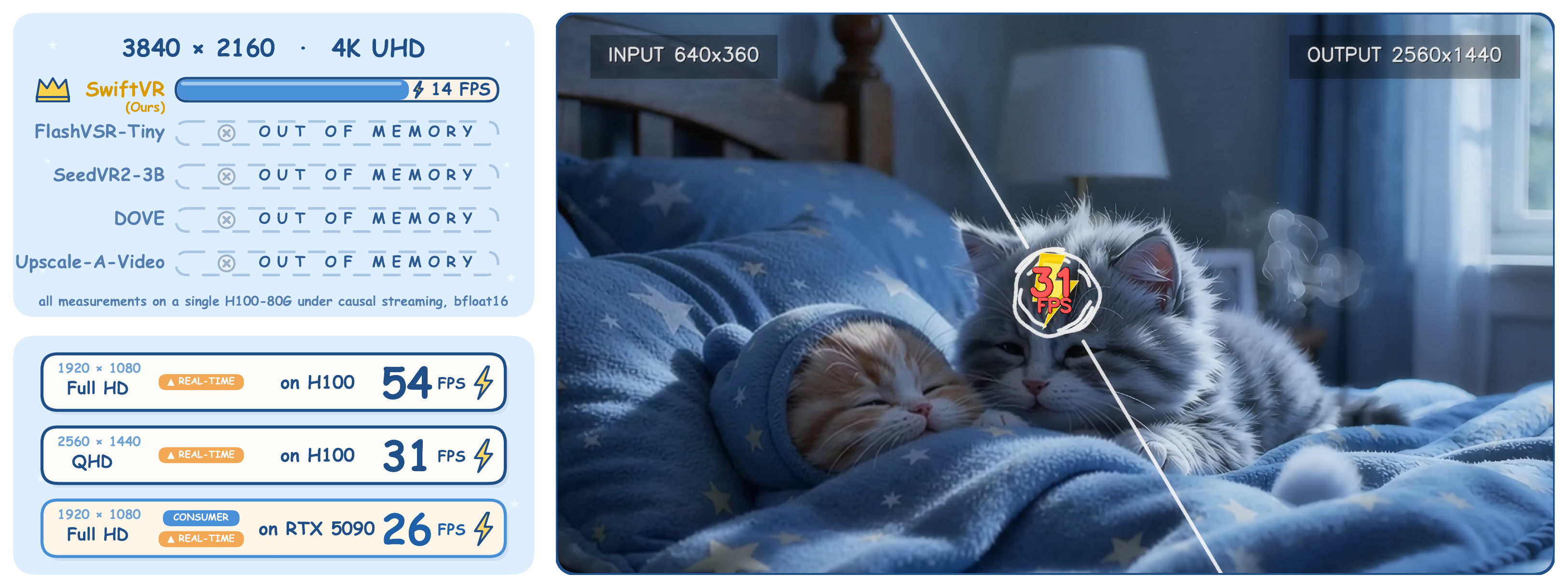}
    \captionof{figure}{SwiftVR enables streaming video restoration at multiple resolutions on a single H100-80G, achieving 54 FPS at Full HD, 31 FPS at QHD (\(2560\times1440\)), and 14 FPS at 4K UHD (\(3840\times2160\)). All compared diffusion-based VR baselines exceed the memory limit at 4K. On a consumer-grade RTX 5090, SwiftVR reaches 26 FPS at 1080p. Right: a \(640\times360\) input is restored to \(2560\times1440\) at 31 FPS.
}

    \label{fig:teaser}
  \end{center}
}]
\thispagestyle{plain}%
\renewcommand\thefootnote{\fnsymbol{footnote}}%
\footnotetext[1]{Equal contribution.}%
\footnotetext[2]{This work was done during Jiaqi Yan's internship at TeleAI.}%
\footnotetext[3]{Corresponding authors: jtzhou@um.edu.mo, xuelong\_li@ieee.org.}%
\makeatother

\begin{abstract}
Real-time video restoration (VR) for live streams requires high-resolution outputs under strict per-frame latency constraints. Existing one-step diffusion-based VR models remain difficult to deploy on consumer-grade GPUs due to two main bottlenecks: quadratic spatial attention at high resolutions and the latency-memory overhead of large video autoencoders. We present SwiftVR, a streaming one-step generative VR framework that reduces both bottlenecks under a causal chunk-wise protocol. For attention, mask-free shifted-window self-attention gathers each spatial window into a dense tensor via deterministic indexing, keeping all attention calls on the dense scaled dot-product attention path without masks, cyclic shifts, padding, or hardware-specific sparse kernels. Because SwiftVR uses only standard dense SDPA calls, the trained model transfers to consumer GPUs without retraining or custom kernels. For autoencoding, a lightweight Restoration-aware Autoencoder enables fast chunk-wise decoding while preserving reconstruction quality. On a single H100, SwiftVR sustains 31~FPS at \(2560\!\times\!1440\) and 14~FPS at \(3840\!\times\!2160\), whereas all compared diffusion-based VR baselines exceed the memory limit at 4K. On a consumer RTX~5090, SwiftVR reaches 26~FPS at \(1920\!\times\!1080\). To our knowledge, SwiftVR is the first generative VR model to achieve real-time 1080p streaming on a consumer-grade GPU, while attaining strong no-reference perceptual quality with lower inference cost.
Project is available at \url{https://h-oliday.github.io/SwiftVR}.
\end{abstract}

\section{Introduction}
\label{sec:intro}

Live video systems increasingly require high-resolution restoration of low-quality streams under strict per-frame latency constraints. A practical system must operate causally, sustain display-resolution throughput, and fit within a consumer-grade GPU memory budget. This remains challenging: real-world video restoration (VR) is severely ill-posed under unknown, time-varying degradations, while streaming precludes offline strategies such as full-clip context and multi-pass refinement.

Prior real-world VR methods fall into three families with distinct quality-efficiency trade-offs. Regression-oriented real-world VR methods~\cite{realesrgan,realbasicvsr,realvsr,realviformer} are efficient and robust to unknown degradations but limited in perceptual realism. Multi-step diffusion methods~\cite{upscaleavideo,venhancer,star} achieve stronger perceptual quality, but repeated sampling incurs prohibitive cost for high-resolution streams. One-step diffusion VR~\cite{dove,seedvr,seedvr2,flashvsr} reduces sampling to a single network evaluation, making streaming-oriented VR more feasible.

\begin{figure}[!h]
  \centering
  \makebox[\linewidth][c]{%
    \includegraphics[width=1.07\linewidth]{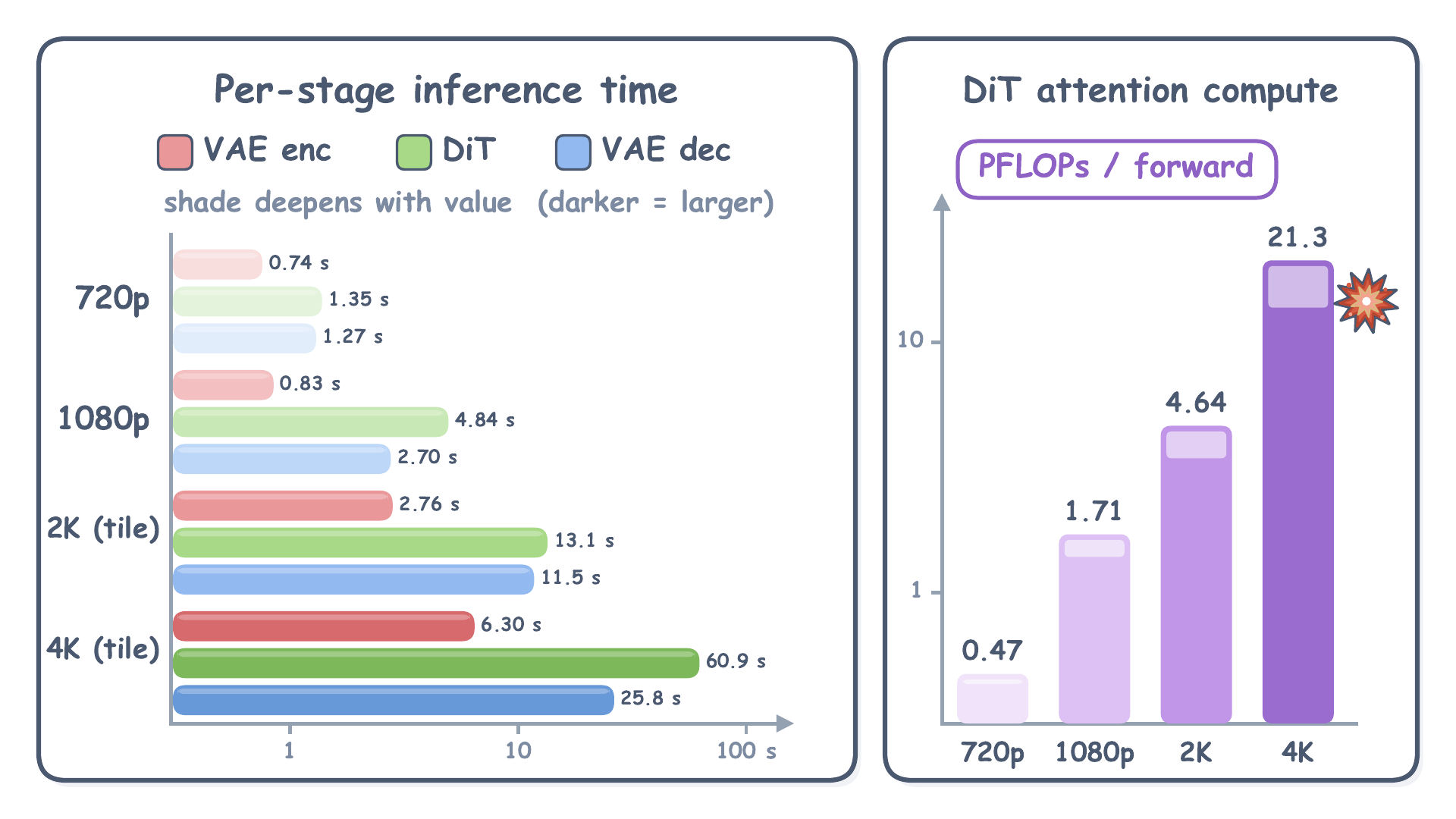}%
  }
  \caption{Latency and attention cost of a single Wan2.2-TI2V-5B forward pass across resolutions on one H100 with bfloat16 and a \(25\)-frame chunk. Left: per-stage inference time; VAE tiling is used at 2K and 4K, and DiT inference dominates at 4K. Right: DiT self-attention computation, increasing from \(0.47\)~PFLOPs at 720p to \(21.3\)~PFLOPs at 4K. Darker shades indicate larger values.}
  \label{fig:intro}
\end{figure}

With one-step sampling, the bottleneck becomes a single high-resolution forward pass. At low resolutions, VAE-DiT generators achieve real-time performance via one-step distillation, KV caching, and sparse attention~\cite{selfforcing,rollingforcing}. However, one-step diffusion VR remains insufficient on consumer hardware at practical restoration resolutions. Because diffusion-based VR uses pretrained video generation backbones, we use Wan2.2-TI2V-5B~\cite{wan} as a representative model to quantify this cost. Even with \((4,16,16)\) VAE compression and \(2\times\) DiT patchification, a single \(3840\!\times\!2160\) forward pass requires \(6.3\)~s, \(60.9\)~s, and \(25.8\)~s for VAE encoding, the DiT, and VAE decoding, respectively, with VAE tiling on one H100 (Figure~\ref{fig:intro}). Two factors dominate this latency. Self-attention over the \(N = THW\) token grid, where \(T\) is temporal length and \(H\!\times\!W\) is spatial size, scales as \(\mathcal{O}(N^2)\); for a fixed aspect ratio, this grows quartically with output width. Once multi-step sampling is removed, encoding and decoding with the 3D VAE also become a substantial part of total latency.

We present SwiftVR, a streaming one-step generative VR framework. It processes streams in causal chunks, bounding temporal extent \(T\) of each DiT tensor and confining quadratic attention growth to spatial axes. This motivates spatial-only rather than general 3D partitioning. Mask-free shifted-window self-attention (MFSWA) gathers each spatial window into a dense tensor, keeping attention calls on the standard scaled dot-product attention (SDPA) fast path. This yields a \(1.62\times\) throughput gain over the full-attention teacher. Unlike Swin attention~\cite{swin}, which uses cyclic shifts and attention masks, MFSWA encodes shifts with deterministic index tensors. Unlike 3D Swin backbones~\cite{seedvr,seedvr2}, which use variable-sized boundary windows, MFSWA handles boundaries via deterministic indexing. This removes operations that would otherwise force SDPA away from the dense path. We introduce a lightweight Restoration-aware Autoencoder (ReAE), jointly fine-tuned with the DiT in pixel space, for fast chunk-wise decoding while preserving reconstruction quality. On a single H100, SwiftVR sustains 31~FPS at \(2560\!\times\!1440\) and 14~FPS at \(3840\!\times\!2160\). Enabled by standard dense SDPA calls, SwiftVR reaches 26~FPS at \(1920\!\times\!1080\) on one RTX~5090 without hardware-specific retraining or kernel rewriting (Figure~\ref{fig:teaser}). In contrast, all compared diffusion-based VR baselines exceed the memory limit at 4K.

In summary, our contributions are threefold. \textbf{(i)} We address real-time generative VR deployment with three designs that reduce attention and autoencoder costs: MFSWA, ReAE, and a causal chunk-wise streaming protocol. Because MFSWA is compatible with standard dense SDPA, the trained model runs across major fused-attention backends and transfers from an H100 to a consumer GPU without retraining or hardware-specific kernels. \textbf{(ii)} We integrate these designs into SwiftVR, a streaming one-step VR model that, to our knowledge, is the first generative VR model to achieve real-time 1080p streaming on a consumer-grade GPU. \textbf{(iii)} Experiments show that SwiftVR achieves leading no-reference perceptual quality among recent one-step VR methods with lower inference cost. It is also the only evaluated diffusion-based method that scales to 4K on a single GPU, where all compared diffusion-based VR baselines exceed the memory limit.

\section{Related Work}
\label{sec:related}

\subsection{Real-world Video Restoration}
Early video restoration methods relied on inter-frame alignment, using motion compensation, deformable convolutions, recurrent propagation, or transformer-based aggregation to exploit temporal redundancy~\cite{basicvsr,basicvsrpp,edvr,vrt,rvrt}. These models typically assume fixed, known degradations, such as bicubic downsampling, and generalize poorly to real-world videos with spatiotemporally varying compression, noise, and blur. Real-world variants therefore use richer synthetic degradation pipelines and introduce cleaning modules to suppress input artifacts before upsampling~\cite{realbasicvsr,realviformer,realvsr}. This prevents residual noise amplification. These methods are efficient and temporally stable, but their regression-oriented objectives optimize pixel-wise errors and bias outputs toward averaged solutions, limiting perceptual realism in heavily degraded regions.

\subsection{One-step Diffusion Video Restoration}
Diffusion priors improve perceptual realism in restoration~\cite{stablesr,diffbir,seesr,supir,venhancer,seedvr,liu2025fape}, but iterative denoising remains prohibitively expensive for high-resolution video streams. Distillation and rectified-flow-based techniques~\cite{lcm,rectifiedflow,dmd,dmd2} compress sampling into a single forward evaluation, and recent studies extend them to VR. DOVE~\cite{dove} fine-tunes a pretrained video diffusion model into a one-step student using a two-stage latent-to-pixel scheme for offline VSR. SeedVR2~\cite{seedvr2} performs one-step VR via diffusion adversarial post-training and adopts adaptive window attention, where the window size is resized according to output resolution. FlashVSR~\cite{flashvsr} formulates streaming VSR as a sparse-attention problem, combining locality-constrained block-sparse attention with a compact decoder. One-step image restoration~\cite{sinsr,osediff,addsr} is computationally efficient per frame but lacks temporal compression and modeling, limiting its extension to efficient and consistent video restoration.

Although these one-step methods substantially reduce sampling cost, they retain bottlenecks that hinder real-time streaming on consumer hardware. These include offline-oriented designs overlooking streaming and autoencoder costs, heavy attention backbones that bottleneck consumer-grade 4K inference, and speedups tied to hardware-specific sparse kernels. FlashVSR reaches \(\sim\!17\)~FPS at \(768\!\times\!1408\) on a server-class A100, remaining below real-time speed and 1080p resolution. Real-time 1080p generative VR on consumer hardware remains unresolved.

\subsection{Efficient Attention in Diffusion}
Once sampling is reduced to a single step, attention computation in the diffusion transformer becomes dominant, motivating three lines of work. The first line is trainable sparse attention~\cite{flashvsr,vsa,spargeattn2}, which achieves high sparsity but relies on dedicated fused sparse kernels for wall-clock speedup. On consumer GPUs without such kernels, sparse arithmetic may not yield measured acceleration. The second line is training-free feature reuse across denoising iterations through caching or forecasting~\cite{deepcache,fastercache,toca,taylorseer,distrifusion}. These methods reduce cost along the sampling axis and thus provide little benefit under single diffusion step inference. The rolling KV cache used by causal streaming generators~\cite{selfforcing,causvid} is orthogonal: it caches prior frames along the temporal axis for cross-chunk consistency rather than reducing per-step attention cost. The third line exploits window-based attention to impose architectural locality, as in SwinIR~\cite{swinir} and Uformer~\cite{uformer}. SeedVR~\cite{seedvr} and SeedVR2~\cite{seedvr2} extend this idea to diffusion transformers using 3D shifted windows, cyclic shifts, attention masks, and variable-sized boundary windows.

Among these alternatives, window-based locality is both kernel-agnostic and compatible with single-evaluation inference, making it suitable for SwiftVR's efficient attention design. However, existing window-based backbones such as SeedVR and SeedVR2 remain offline-oriented and rely on 3D shifted windows, cyclic shifts, and attention masks to process full-sequence inputs at arbitrary resolutions. In SwiftVR, the temporal extent is already bounded by the chunk length, so window partitioning is applied only along the spatial dimensions. Cross-window information is exchanged through alternating non-shifted and half-shifted spatial layouts, without cyclic shifts or attention masks in the training graph. This design keeps all attention operations compatible with standard dense SDPA kernels, avoiding custom sparse kernels and mask-induced fallback paths.

\begin{figure*}[t]
    \centering
    \includegraphics[width=\linewidth]{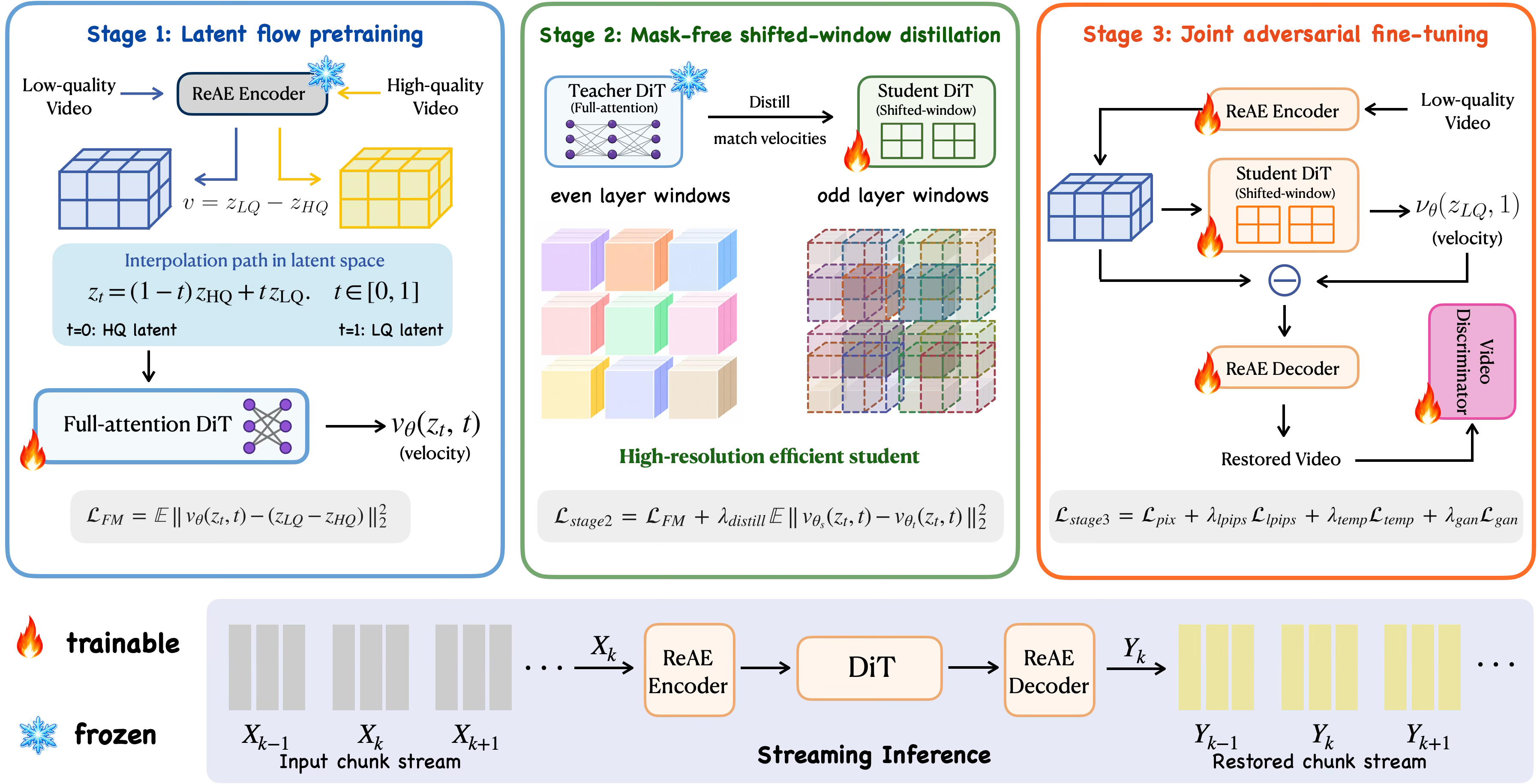}
\caption{%
\textbf{Overview of the SwiftVR pipeline.} SwiftVR optimizes the DiT in three stages and performs causal streaming inference.
\textbf{(a) Stage 1:} In the ReAE latent space, a full-attention DiT learns the constant velocity \(v = z_\mathrm{LQ} - z_\mathrm{HQ}\) along \(z_t = (1{-}t)z_\mathrm{HQ} + t z_\mathrm{LQ}\).
\textbf{(b) Stage 2:} The full-attention teacher is distilled into a shifted-window student that partitions only the spatial axes and alternates non-shifted with half-window-shifted layouts, preserving dense tensors within each window.
\textbf{(c) Stage 3:} The DiT, ReAE, and video discriminator are jointly fine-tuned under the deployment-time one-step inference protocol.
\textbf{(d) Streaming inference:} With all modules frozen, each input chunk \(X_k\) is restored to \(Y_k\) using a single DiT pass.
The fire and snowflake icons indicate trainable and frozen modules, respectively.}
\label{fig:pipeline}

\end{figure*}

\section{Method}
\label{sec:method}

SwiftVR is a streaming, one-step generative video restoration framework comprising a compact autoencoder and a window-based self-attention diffusion transformer. SwiftVR processes videos causally in fixed-size chunks, thereby bounding the temporal length \(T\) of each DiT tensor. Because self-attention scales quadratically with \(N=T H W\), where \(H\) and \(W\) denote latent spatial height and width, chunking limits temporal growth and motivates spatial-only rather than full 3D window partitioning.
The diffusion transformer is optimized in three stages: full-attention latent training, mask-free shifted-window distillation, and joint pixel-space fine-tuning with the ReAE. At inference, SwiftVR restores the input stream chunk by chunk under the same causal protocol. Figure~\ref{fig:pipeline} illustrates the DiT optimization stages and the streaming inference pipeline.

\subsection{Restoration-aware Autoencoder}
\label{sec:reae}

As one-step generative restoration reduces the sampling cost, the autoencoder emerges as a major source of end-to-end latency. The original 3D VAE used in large video generation backbones~\cite{wan} incurs high latency for real-time high-resolution decoding and is difficult to jointly optimize with the DiT. We therefore introduce ReAE, a compact restoration-aware autoencoder serving as the latent interface. ReAE is initialized from a publicly available lightweight autoencoder~\cite{tae} and adapted to video restoration through fine-tuning on video data.

ReAE is trained independently on clean videos in two stages. The first stage optimizes pixel fidelity, perceptual similarity, and temporal consistency:
\begin{equation}
\mathcal{L}_\text{ReAE}^{(1)}
= \mathcal{L}_\text{pix}
+ \lambda_\text{lpips}^{\text{ReAE}}\mathcal{L}_\text{lpips}
+ \lambda_\text{temp}^{\text{ReAE}}\mathcal{L}_\text{temp},
\end{equation}
where \(\mathcal{L}_\text{pix}=\|\hat{x}-x\|_1\) and \(\mathcal{L}_\text{temp}\) denotes the MSE between consecutive frame differences. The second stage adds adversarial supervision after reconstruction training converges:
\begin{equation}
\mathcal{L}_\text{ReAE}^{(2)}
= \mathcal{L}_\text{ReAE}^{(1)}
+ \lambda_\text{gan}^{\text{ReAE}}\mathcal{L}_\text{gan}.
\end{equation}
ReAE is frozen during latent flow matching and updated during joint pixel-space fine-tuning with the DiT.

\subsection{Progressive DiT Optimization}
\label{sec:dit}

\paragraph{Stage 1: Full-attention latent flow matching.}
We train a full-attention DiT in the frozen ReAE latent space to predict the displacement from a low-quality latent video to its high-quality counterpart. We encode the high- and low-quality videos as \(z_\text{HQ}\!=\!E_\phi(x_\text{HQ})\) and \(z_\text{LQ}\!=\!E_\phi(x_\text{LQ})\), respectively. We then define the linear path \(z_t = (1-t) z_\text{HQ} + t\,z_\text{LQ}\), \(t \in [0,1]\), with constant velocity \(z_\text{LQ} - z_\text{HQ}\). We place the high-quality endpoint at \(t\!=\!0\) and the low-quality endpoint at \(t\!=\!1\), enabling a single backward step from the inference-time input to recover the high-quality latent. The DiT is trained to predict this constant degradation velocity:
\begin{equation}
\mathcal{L}_\text{FM} = \mathbb{E}_{z_\text{HQ}, z_\text{LQ}, t}\!\left[ \big\| v_\theta(z_t, t) - (z_\text{LQ} - z_\text{HQ}) \big\|_2^2 \right].
\label{eq:fm}
\end{equation}

Uniform sampling of \(t\) provides mixed-level latent augmentation and encourages the network to estimate a \(t\)-invariant displacement across interpolation levels.

\paragraph{Stage 2: Mask-free shifted-window distillation.}

With \(N\!\approx\!10^5\) tokens, self-attention accounts for over \(60\%\) of the Stage-1 DiT latency. Although a block-diagonal mask reduces the nominal attention range, it often disables fused dense SDPA backends and triggers fallback to materialized attention~\cite{flashattn2,sageattn,xformers}. We therefore encode the window structure outside the attention kernel using deterministic gather and scatter operations.

We introduce mask-free shifted-window self-attention (MFSWA), which invokes attention through the standard scaled dot-product interface with \texttt{attn\_mask=None} and no padding tokens. Unlike Swin SW-MSA~\cite{swin}, which implements shifted windows using cyclic shifts and attention masks, MFSWA realizes shifts through deterministic priority-coherent scatter. Unlike SeedVR and SeedVR2~\cite{seedvr,seedvr2}, which handle varying or non-divisible resolutions by resizing windows or introducing variable-sized boundary windows, MFSWA retains a fixed window size. Boundary cases are handled by uniform-shape boundary-clamped gather without per-resolution geometry changes. Together, these design choices remove the operations that would otherwise force SDPA off the dense path.

MFSWA is defined by three core design choices, as shown in Fig.~\ref{fig:swa}; details of boundary-clamped gather are provided in the supplementary material. (i) Spatial-only partition: partitioning is applied only over \((H,W)\), while all \(T\) frames in a chunk remain jointly visible within each window. (ii) Dense-block pre-gather: each window is gathered into a dense tensor, enabling per-window attention through a single dense SDPA call on \(Q,K,V\). (iii) Half-window shift with priority-coherent scatter: even layers use non-shifted windows, whereas odd layers apply a half-window shift \((w_h/2, w_w/2)\). Each output token is assigned to a deterministic owner window, enabling cross-window information flow without cyclic shifts or masks.

The student is trained with flow matching and an additional teacher-distillation term:
\begin{equation}
\small
\mathcal{L}_\text{stage2} = \mathcal{L}_\text{FM} + \lambda_\text{distill}\, \mathbb{E}_{z_t, t}\!\left[ \big\| v_{\theta_s}(z_t,t) - v_{\theta_t}(z_t,t) \big\|_2^2 \right].
\label{eq:distill}
\end{equation}

\begin{figure}[t]
    \centering
    \includegraphics[width=\linewidth]{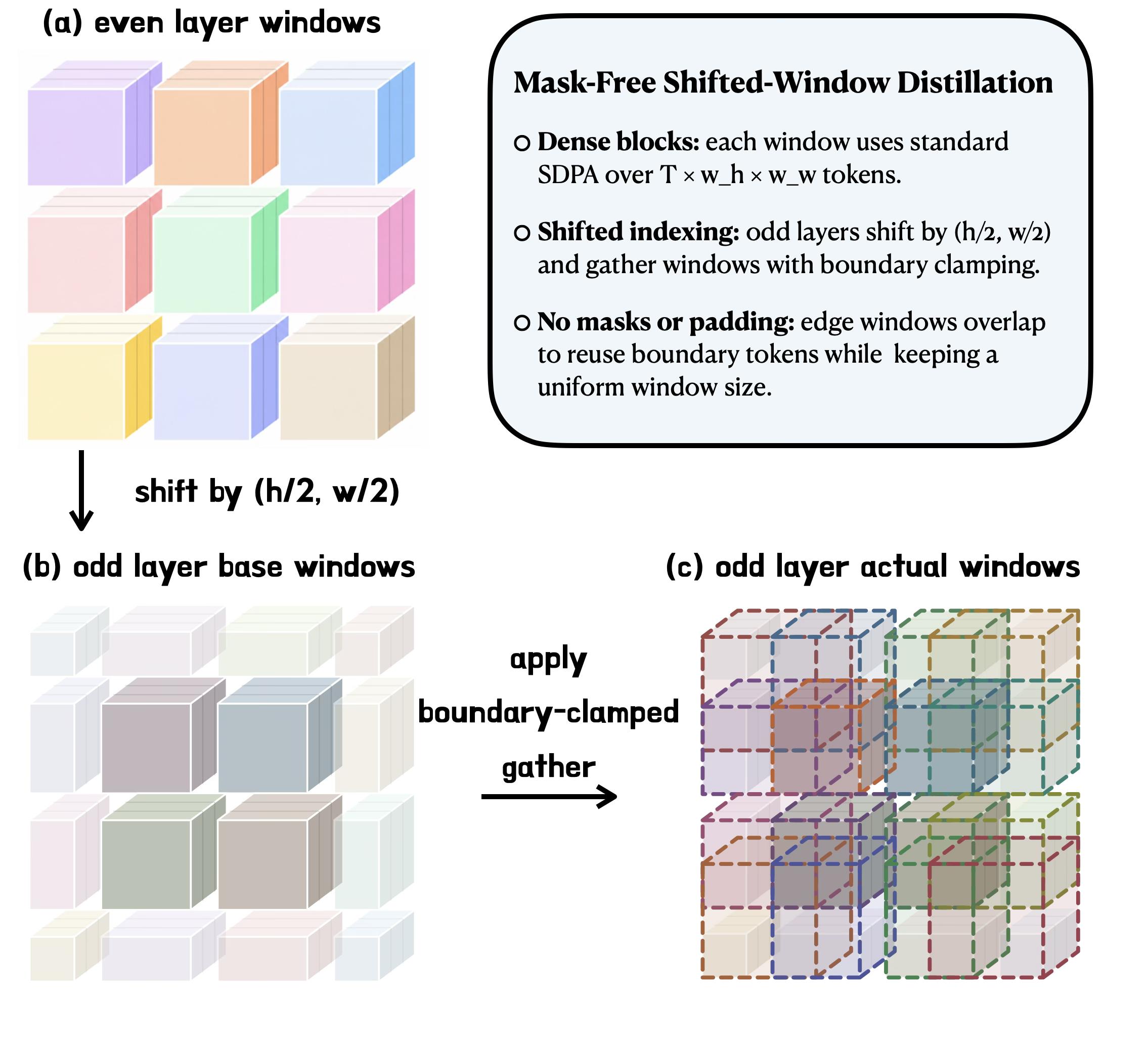}
    \caption{Illustration of mask-free shifted-window attention. 
(a) Even-layer windows. 
(b) Half-window-shifted base partition. 
(c) Odd-layer effective windows, shown as dashed cubes; each is pre-gathered into a dense tensor and processed by standard scaled dot-product attention without masks, cyclic shifts, or padding.
}
    \label{fig:swa}
\end{figure}

\paragraph{Stage 3: Joint adversarial fine-tuning.}
The flow-matching objective in the previous stages is defined entirely in latent space and therefore constrains the decoded pixel-space output only indirectly. To close this latent-to-pixel gap, we jointly fine-tune the DiT and ReAE under the deployment-time one-step inference protocol. Starting from \(t\!=\!1\), the model subtracts the predicted velocity in a single forward pass and decodes the resulting latent:
\begin{equation}
\small
\hat{z}_\text{HQ} = E_\phi(x_\text{LQ}) - v_\theta\!\big(E_\phi(x_\text{LQ}),\,1\big), \quad \hat{x} = D_\phi(\hat{z}_\text{HQ}).
\label{eq:onestep}
\end{equation}
The decoded output is supervised in pixel space using \(\mathcal{L}_\text{stage3} = \mathcal{L}_\text{pix} + \lambda_\text{lpips}^{\text{S3}}\mathcal{L}_\text{lpips} + \lambda_\text{temp}^{\text{S3}}\mathcal{L}_\text{temp} + \lambda_\text{gan}^{\text{S3}}\mathcal{L}_\text{gan}\). For adversarial supervision, we employ a video discriminator based on a frozen VGG-19 backbone~\cite{vgg}. Specifically, we extract frame-wise multi-scale perceptual features, reorganize them into spatio-temporal feature volumes, and feed them into trainable spectral-normalized 3D patch heads. The resulting multi-scale video patch logits promote sharper perceptual details while suppressing temporally inconsistent artifacts.

\subsection{Streaming Inference}
\label{sec:streaming}
We adopt a causal chunk-wise streaming protocol. The stream is divided into \(L\)-frame non-overlapping chunks aligned with the temporal stride of the streaming ReAE. Each DiT forward pass processes only the latent chunk, without access to future frames, overlapped inference, or a rolling KV cache. Cross-chunk continuity is handled by the streaming ReAE, which maintains encoder and decoder boundary states across chunks.

For the first chunk, the decoder discards causal-padding frames; middle chunks are emitted directly; for the last chunk, the input is padded to satisfy the ReAE stride, and only valid frames are retained. Here, \(s_E^k\) and \(s_D^k\) denote encoder and decoder boundary states after chunk \(k\), respectively. Formally, the streaming encoder produces \(z^k_\mathrm{LQ},\ s_E^k = E_\phi^{\mathrm{str}}(X_k,\ s_E^{k-1})\). The DiT then predicts one-step velocity independently for each chunk:

\begin{equation}
\small
\hat{z}^k_\mathrm{HQ} = z^k_\mathrm{LQ} - v_\theta(z^k_\mathrm{LQ},\ 1),\quad
\hat{X}_k,\ s_D^k = D_\phi^{\mathrm{str}}(\hat{z}^k_\mathrm{HQ},\ s_D^{k-1}).
\label{eq:stream}
\end{equation}


\begin{table*}[htbp]
\centering
\caption{Quantitative comparison on synthetic and real-world video restoration benchmarks. Methods are grouped by method type for comparison. $\uparrow$ and $\downarrow$ indicate that higher and lower values are better, respectively. The best and second-best results are highlighted in red.}
\label{tab:main}
\setlength{\tabcolsep}{2.5pt}
\renewcommand{\arraystretch}{1.0}
\resizebox{\textwidth}{!}{
\begin{tabular}{c|l|ccc|ccccc}
\toprule
Dataset & Metric 
& Real-ESRGAN~\cite{realesrgan}
& RealBasicVSR~\cite{realbasicvsr}
& RealViFormer~\cite{realviformer}
& UAV~\cite{upscaleavideo}
& DOVE~\cite{dove}
& SeedVR2-3B~\cite{seedvr2}
& FlashVSR-Tiny~\cite{flashvsr}
& SwiftVR(Ours) \\
\midrule
\multirow{8}{*}{SPMCS~\cite{spmc}}
& PSNR~$\uparrow$  & 21.67 & 22.54 & 22.96 & 20.91 & \textcolor{red}{23.31} & 22.23 & 21.56 & 22.33 \\
& SSIM~$\uparrow$  & 0.5686 & 0.5743 & 0.5958 & 0.4795 & \textcolor{red}{0.6272} & 0.5984 & 0.5466 & 0.5742 \\
& LPIPS~$\downarrow$ & 0.3631 & 0.3572 & 0.3288 & 0.4097 & 0.2745 & \textcolor{red}{0.2619} & 0.2736 & 0.2837 \\
& DISTS~$\downarrow$ & 0.2204 & 0.2121 & 0.2083 & 0.2361 & 0.1662 & \textcolor{red}{0.1410} & 0.1631 & 0.1535 \\
& CLIP-IQA~$\uparrow$ & 0.5079 & 0.4354 & 0.3994 & \textcolor{red}{0.5712} & 0.5127 & 0.5441 & 0.5314 & 0.5011 \\
& MUSIQ~$\uparrow$ & 66.47 & 63.29 & 64.63 & 69.12 & 69.67 & 68.76 & 69.94 & \textcolor{red}{71.74} \\
& MANIQA~$\uparrow$ & 0.3938 & 0.3350 & 0.3116 & 0.4067 & 0.3871 & \textcolor{red}{0.4137} & 0.3961 & 0.3866 \\
& NIQE~$\downarrow$ & 3.3081 & 3.3123 & 3.5901 & \textcolor{red}{3.1558} & 4.617 & 3.5371 & 3.6077 & 3.4166 \\
\midrule
\multirow{8}{*}{UDM10~\cite{udm10}}
& PSNR~$\uparrow$  & 24.52 & 25.09 & 26.24 & 24.57 & \textcolor{red}{26.97} & 26.13 & 24.02 & 25.58 \\
& SSIM~$\uparrow$  & 0.7313 & 0.7331 & 0.7690 & 0.6883 & \textcolor{red}{0.7953} & 0.7712 & 0.7028 & 0.7659 \\
& LPIPS~$\downarrow$ & 0.3281 & 0.3291 & 0.2885 & 0.3249 & \textcolor{red}{0.2172} & 0.2192 & 0.2483 & 0.2508 \\
& DISTS~$\downarrow$ & 0.1901 & 0.1868 & 0.1752 & 0.1774 & 0.1291 & \textcolor{red}{0.1089} & 0.1334 & 0.1311 \\
& CLIP-IQA~$\uparrow$ & 0.4611 & 0.4482 & 0.4022 & 0.4355 & 0.4744 & 0.4032 & 0.4733 & \textcolor{red}{0.5241} \\
& MUSIQ~$\uparrow$ & 58.31 & 62.03 & 58.85 & 60.60 & 63.33 & 57.52 & 65.95 & \textcolor{red}{67.34} \\
& MANIQA~$\uparrow$ & 0.3398 & 0.3194 & 0.2929 & 0.2907 & 0.3423 & 0.2832 & 0.3588 & \textcolor{red}{0.3609} \\
& NIQE~$\downarrow$ & 4.0600 & \textcolor{red}{3.6600} & 4.0476 & 3.9552 & 5.0568 & 4.5434 & 3.7783 & 3.8798 \\
\midrule
\multirow{8}{*}{YouHQ40~\cite{upscaleavideo}}
& PSNR~$\uparrow$  & 22.75 & 22.35 & 23.44 & 22.17 & \textcolor{red}{24.29} & 23.11 & 22.24 & 22.67 \\
& SSIM~$\uparrow$  & 0.6336 & 0.5873 & 0.6234 & 0.5651 & \textcolor{red}{0.6722} & 0.6427 & 0.5871 & 0.6048 \\
& LPIPS~$\downarrow$ & 0.3642 & 0.4137 & 0.3773 & 0.3781 & 0.3039 & \textcolor{red}{0.2924} & 0.2989 & 0.2947 \\
& DISTS~$\downarrow$ & 0.1844 & 0.1979 & 0.1982 & 0.1862 & 0.1485 & 0.1265 & 0.1373 & \textcolor{red}{0.1195} \\
& CLIP-IQA~$\uparrow$ & 0.4518 & 0.4903 & 0.4410 & 0.4876 & 0.4551 & 0.4845 & 0.5227 & \textcolor{red}{0.5950} \\
& MUSIQ~$\uparrow$ & 57.28 & 64.72 & 62.30 & 60.77 & 60.86 & 60.46 & 66.98 & \textcolor{red}{69.81} \\
& MANIQA~$\uparrow$ & 0.3078 & 0.3012 & 0.2841 & 0.3040 & 0.3061 & 0.3227 & 0.3659 & \textcolor{red}{0.3669} \\
& NIQE~$\downarrow$ & 3.8349 & \textcolor{red}{3.1222} & 3.3621 & 3.4593 & 4.8584 & 3.5318 & 3.4382 & 3.1634 \\
\midrule
\multirow{4}{*}{VideoLQ~\cite{realbasicvsr}}
& CLIP-IQA~$\uparrow$ & 0.3619 & \textcolor{red}{0.3801} & 0.3465 & 0.2892 & 0.2880 & 0.2319 & 0.3616 & 0.3789 \\
& MUSIQ~$\uparrow$ & 49.86 & 55.05 & 52.12 & 45.23 & 44.59 & 40.42 & 50.93 & \textcolor{red}{55.43} \\
& MANIQA~$\uparrow$ & 0.2903 & \textcolor{red}{0.3006} & 0.2796 & 0.2416 & 0.2566 & 0.2179 & 0.2825 & 0.2767 \\
& NIQE~$\downarrow$ & 4.2017 & \textcolor{red}{3.7051} & 4.0577 & 4.6649 & 5.3566 & 5.3106 & 4.0159 & 4.0468 \\
\bottomrule
\end{tabular}
}
\end{table*}

\section{Experiments}
\label{sec:experiments}

\subsection{Experimental Setup}

\paragraph{Implementation.} SwiftVR is based on Wan2.2-TI2V-5B~\cite{wan} and trained with AdamW~\cite{adamw} and DeepSpeed ZeRO-2~\cite{zero} on \(8\!\times\!\) H100-80G GPUs. We use 33-frame \(768\!\times\!1280\) clips for ReAE pretraining, latent flow matching, and window-attention distillation, and 13-frame multi-resolution clips for joint fine-tuning. The learning rate is set to \(2\!\times\!10^{-5}\) for Stage~1 and \(1\!\times\!10^{-5}\) for Stages~2 and~3.
For ReAE training, the loss weights are
\(\lambda_\text{lpips}^{\text{ReAE}}=1.0\),
\(\lambda_\text{temp}^{\text{ReAE}}=1.0\), and
\(\lambda_\text{gan}^{\text{ReAE}}=0.05\).
For joint fine-tuning, we use
\(\lambda_\text{lpips}^{\text{S3}}=0.5\),
\(\lambda_\text{temp}^{\text{S3}}=1.0\), and
\(\lambda_\text{gan}^{\text{S3}}=1.0\).
The distillation weight is set to
\(\lambda_\text{distill}=1.0\).
All stages are trained on curated high-quality clips from UltraVideo~\cite{ultravideo}. Paired low- and high-quality videos are synthesized using the RealBasicVSR degradation pipeline~\cite{realbasicvsr}.

\paragraph{Evaluation.}
We evaluate on three synthetic benchmarks: SPMCS~\cite{spmc}, UDM10~\cite{udm10}, and YouHQ40~\cite{upscaleavideo}. These benchmarks use the same degradation protocol as training. We also evaluate on the real-world VideoLQ benchmark~\cite{realbasicvsr}. All methods are evaluated under a unified chunk-based streaming protocol, with implementation details provided in the supplementary material. We compare SwiftVR with three categories of real-world video restoration baselines: non-diffusion methods, including Real-ESRGAN~\cite{realesrgan}, RealBasicVSR~\cite{realbasicvsr}, and RealViFormer~\cite{realviformer}; the multi-step diffusion method Upscale-A-Video~\cite{upscaleavideo}; and one-step diffusion methods, including DOVE~\cite{dove}, SeedVR2-3B~\cite{seedvr2}, and FlashVSR-Tiny~\cite{flashvsr}.


\paragraph{Metrics.} For full-reference synthetic benchmarks, we report PSNR/SSIM for fidelity, LPIPS/DISTS for perceptual similarity, and CLIP-IQA, MUSIQ, MANIQA, and NIQE as no-reference metrics. For real-world benchmarks, we report only no-reference metrics. For streaming deployment, we report FPS and peak GPU memory.

\subsection{Comparison with Existing Methods}

\paragraph{Quantitative Comparisons.}
Table~\ref{tab:main} summarizes the quantitative results. SwiftVR consistently achieves strong perceptual quality across all benchmarks, especially on no-reference metrics. It ranks first in MUSIQ on all four benchmarks and first in CLIP-IQA and MANIQA on UDM10 and YouHQ40. For DISTS, a full-reference perceptual metric, SwiftVR ranks first on YouHQ40 and second on SPMCS. On LPIPS, SwiftVR remains competitive, trailing the leading one-step method by only a small margin. Fidelity-oriented methods such as DOVE obtain higher PSNR and SSIM because their objectives emphasize pixel accuracy rather than perceptual detail. Because fidelity and perceptual realism often favor different restoration behaviors, SwiftVR prioritizes perceptual quality, which is more aligned with real-world video restoration.

\begin{figure*}[!t]
    \centering
    \includegraphics[width=\linewidth]{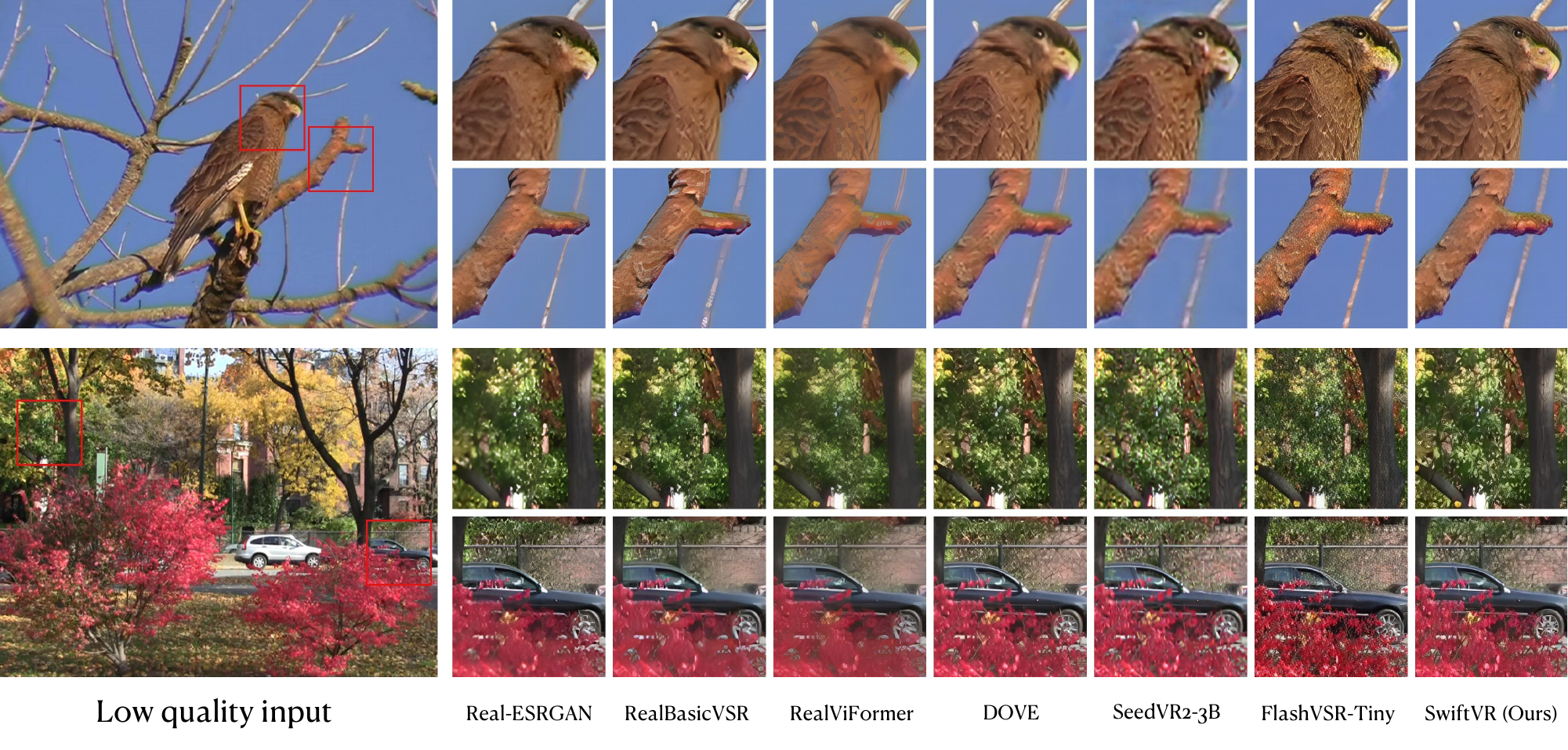}
    \caption{Qualitative comparison on real-world video clips. \textbf{Top:} a perched falcon, with crops showing the head, beak, and a bare branch against the sky. \textbf{Bottom:} a residential street with autumn foliage and a parked car, including crops of the dense leaf canopy and the car behind a chain-link fence. Columns from left to right show the low-quality input (LQ), Real-ESRGAN~\cite{realesrgan}, RealBasicVSR~\cite{realbasicvsr}, RealViFormer~\cite{realviformer}, DOVE~\cite{dove}, SeedVR2-3B~\cite{seedvr2}, FlashVSR-Tiny~\cite{flashvsr}, and SwiftVR (Ours). Best viewed at high magnification.}
    \label{fig:visual_fig}
\end{figure*}

\paragraph{Qualitative Comparisons.}
Figure~\ref{fig:visual_fig} compares visual quality on real-world videos using enlarged local patches. The two examples cover complementary restoration challenges: fine feather textures on the falcon's head and beak, and repeated thin structures in the street scene, including branches, foliage, fences, and the car. Regression-based baselines, including Real-ESRGAN, RealBasicVSR, and RealViFormer, recover global silhouettes but oversmooth fine details and introduce color fringing along branches. Although DOVE achieves higher PSNR and SSIM, its outputs exhibit over-smoothed head feathers and foliage, reflecting its stronger emphasis on pixel fidelity. SeedVR2-3B and FlashVSR-Tiny recover more high-frequency content but introduce localized color shifts, halos, or over-sharpening near branches and car contours. In contrast, SwiftVR produces sharper and more natural reconstructions, with directional feather textures, cleaner beak details, clearer branch boundaries, better leaf separation, and sharper car contours. These observations are consistent with the improvements in perceptual metrics. RealBasicVSR performs slightly better on VideoLQ no-reference metrics, but its visual results remain overly smooth.

\begin{table}[!h]
\centering
\caption{Efficiency comparison of one-step video restoration methods at \(2560\!\times\!1440\) on a single H100 under causal streaming, measured over \(24\) output frames. DOVE and SeedVR2-3B exceed the memory limit with their default VAEs
; therefore, we enable \texttt{use\_tile=True}.}
\label{tab:portability}
\setlength{\tabcolsep}{2pt}
\resizebox{\linewidth}{!}{
\begin{tabular}{l|cccc}
\toprule
Metric & DOVE~\cite{dove} & SeedVR2-3B~\cite{seedvr2} & FlashVSR-Tiny~\cite{flashvsr} & SwiftVR (Ours) \\
\midrule
Avg. Time (s) & 27.615 & 17.320 & 2.493 & 0.766 \\
FPS & 0.87 & 1.39 & 9.61 & 31.32 \\
Peak Mem. (GB) & 59.24 & 35.35 & 34.35 & 38.01 \\
\bottomrule
\end{tabular}}
\end{table}

\subsection{Ablation Study}

\begin{table}[!h]
\centering
\caption{Ablation study of the MFSWA design. Masked SWA replaces dense pre-gathering with a block-diagonal attention mask, which disables the fused SDPA execution path.}
\label{tab:ablation_mfsw}
\setlength{\tabcolsep}{3pt}
\resizebox{\linewidth}{!}{
\begin{tabular}{lccccc}
\toprule
Variant & PSNR$\uparrow$ & LPIPS$\downarrow$ & DiT Time (ms) & Peak Mem. (GB) & FPS$\uparrow$ \\
\midrule
Full Attention (Teacher) & 25.86 & 0.2417 & 1039.11 & 35.37 & 19.36 \\
Masked SWA               & 25.34 & 0.2637 & 674.49 & 38.17 & 27.47 \\
MFSWA (Ours)             & 25.58 & 0.2508 & 566.31 & 38.01 & \textbf{31.32} \\
\bottomrule
\end{tabular}
}
\end{table}

\paragraph{Mask-free shifted-window self-attention (MFSWA).} We compare three self-attention variants on the same backbone using window size \((w_h, w_w)\!=\!(16, 16)\), bfloat16, and a \(2560\!\times\!1440\) causal-streaming protocol (Table~\ref{tab:ablation_mfsw}). They are: (1) the full-attention teacher; (2) a Masked SWA student using the same spatial-only window partition but implementing it with a block-diagonal SDPA mask; and (3) our MFSWA student. Variants (2) and (3) use identical window geometry and training settings; their only difference is masked attention versus dense pre-gathering. This highlights the advantage of encoding window structure outside the attention kernel. Although Masked SWA uses the same spatial partition as MFSWA, its block-diagonal mask disables fused Flash/cuDNN SDPA backends and triggers fallback to a materialized attention path. As a result, it improves over the full-attention teacher but remains slower than MFSWA (\(27.47\) vs.\ \(31.32\)~FPS) and incurs the highest peak memory (\(38.17\)~GB). In contrast, MFSWA keeps each window as a dense SDPA input, converting the window partition into a practical speedup. It reaches \(31.32\)~FPS, \(1.62\times\) teacher throughput, while maintaining comparable restoration quality (\(25.58\) vs.\ \(25.86\)~dB PSNR; \(0.2508\) vs.\ \(0.2417\) LPIPS). These results demonstrate that MFSWA benefits not only from local attention but also from mask-free implementation, which keeps attention calls on the efficient dense path.


\begin{table}[!h]
\centering
\caption{Ablation study of ReAE on \(25\!\times\!1088\!\times\!1920\) videos.}
\label{tab:ablation_reae}
\setlength{\tabcolsep}{4pt}
\resizebox{\linewidth}{!}{
\begin{tabular}{lcccccc}
\toprule
Variant & PSNR$\uparrow$ & LPIPS$\downarrow$ & Params (M) &  Peak Mem. (GB) $\downarrow$ & Enc. Time (s) & Dec. Time (s) \\
\midrule
Wan2.2-VAE   & 35.48 & 0.0513 & 704.69 & 24.86 & 0.818 & 2.714 \\ 
Tiny autoencoder   & 27.14 & 0.1183 & 11.42 & 7.74 & 0.033 & 0.040 \\
ReAE              & 32.74 & 0.0777 & 40.95 & 16.97 & 0.034 & 0.099 \\
\bottomrule
\end{tabular}

}
\end{table}

\paragraph{ReAE.} To assess the autoencoder design, we compare ReAE with the original Wan2.2-VAE~\cite{wan} and a generic tiny autoencoder~\cite{tae} on \(25\!\times\!1088\!\times\!1920\) clips (Table~\ref{tab:ablation_reae}). The original Wan2.2-VAE attains the best reconstruction quality (\(35.48\)~dB PSNR, \(0.0513\) LPIPS). However, it is also the most expensive, requiring \(704.69\)M parameters, \(24.86\)~GB peak memory, and \(2.714\)s decoding per chunk. The tiny autoencoder is lightweight (\(11.42\)M parameters and \(0.040\)s decoding) but has lower reconstruction quality (\(27.14\)~dB PSNR, \(0.1183\) LPIPS). ReAE achieves a stronger quality-efficiency trade-off, with \(40.95\)M parameters, \(0.034\)s encoding time, \(0.099\)s decoding time, and \(16.97\)~GB peak memory. It also improves reconstruction quality to \(32.74\)~dB PSNR and \(0.0777\) LPIPS, substantially outperforming the tiny autoencoder.
These results show that ReAE substantially alleviates the autoencoder bottleneck and makes joint fine-tuning with the DiT tractable.

\begin{table}[!h]
\centering
\caption{Runtime breakdown of SwiftVR on a single H100 using bfloat16 and the default streaming protocol.}
\label{tab:runtime_breakdown}
\setlength{\tabcolsep}{3pt}
\resizebox{\linewidth}{!}{
\begin{tabular}{lccccc}
\toprule
Resolution & Enc. (ms) $\downarrow$ & DiT (ms) $\downarrow$ & Dec. (ms) $\downarrow$ & Peak Mem. (GB) $\downarrow$ & FPS $\uparrow$\\
\midrule
1920$\times$1080 & 25.67  & 327.72 & 85.37 & 29.26 & 54.42 \\
2560$\times$1440 & 45.10 & 566.31 & 151.74 & 38.01 & 31.32 \\
3840$\times$2160 & 111.06 & 1270.10 & 344.27 & 60.91 & 13.84 \\
\bottomrule
\end{tabular}
}
\end{table}

\subsection{Efficiency Analysis}

At \(2560\!\times\!1440\), SwiftVR is the most efficient one-step diffusion video restoration method (Table~\ref{tab:portability}). It reaches \(31.32\)~FPS, corresponding to \(0.766\)~s per \(24\)-frame chunk. This is approximately \(3.3\times\) the throughput of FlashVSR-Tiny and an order of magnitude higher than DOVE and SeedVR2-3B, which fit this resolution only with VAE tiling.

This advantage increases at higher resolutions. At \(3840\!\times\!2160\), all compared diffusion-based VR baselines exceed the memory limit on a single H100, whereas SwiftVR sustains \(13.84\)~FPS, making it the only evaluated method capable of 4K inference on a single GPU. The per-component breakdown in Table~\ref{tab:runtime_breakdown} shows that the DiT dominates end-to-end latency across resolutions. This is consistent with one-step video restoration shifting the bottleneck from iterative sampling to the per-step transformer computation.

Compared with the full-attention teacher using the same backbone
(\(19.36\!\to\!31.32\)~FPS, Table~\ref{tab:ablation_mfsw}),
MFSWA replaces a full \(T H W\)-token attention call with multiple dense local attention calls of length \(T w_h w_w\), while keeping attention calls on the dense-attention fast path.

For consumer-grade deployment, we benchmark SwiftVR on a single NVIDIA RTX~5090 at \(1920\!\times\!1080\) under the same chunk protocol.
SwiftVR sustains \(\!26\)~FPS with default chunk length \(L\!=\!24\), within the \(24\)--\(30\)~FPS budget for live streaming, video conferencing, and cloud gaming. To our knowledge, SwiftVR is the first generative video restoration model to achieve real-time 1080p streaming on a consumer-grade GPU. A closely related one-step streaming diffusion VSR method, FlashVSR~\cite{flashvsr}, reports \(17\)~FPS at \(768\!\times\!1408\) on a server-class A100. It relies on block-sparse acceleration based on a FlashAttention-2 kernel, whose availability depends on GPU architecture.
In contrast, MFSWA uses standard dense SDPA calls, allowing SwiftVR to transfer to the RTX~5090 without hardware-specific retraining or kernel rewriting.

\section{Conclusion}

We present SwiftVR, a one-step generative framework for real-time video restoration. To our knowledge, SwiftVR is the first generative method to achieve real-time 1080p streaming video restoration on a consumer-grade GPU. It restores low-quality streams with a causal chunk-wise protocol and addresses the two dominant costs of one-step diffusion VR through complementary attention and autoencoder designs. Mask-free shifted-window self-attention confines attention to fixed-size spatial windows while preserving standard dense SDPA execution, achieving a \(1.62\times\) speedup over the full-attention teacher without hardware-specific retraining or kernel rewriting. The lightweight restoration-aware autoencoder further reduces decoding cost while preserving reconstruction quality.

Experiments show that SwiftVR attains strong no-reference perceptual quality among one-step VR methods with lower inference cost. On a single H100, SwiftVR sustains \(31\)~FPS at \(2560\!\times\!1440\) and \(14\)~FPS at \(3840\!\times\!2160\), making it the only evaluated diffusion-based VR method supporting 4K inference on a single GPU; compared diffusion-based VR baselines exceed the memory limit at 4K. On a consumer RTX~5090, SwiftVR reaches \(26\)~FPS at \(1920\!\times\!1080\). Real-time generative 4K restoration on consumer hardware remains an open challenge and motivates future work on inference acceleration and compact backbones.

\newpage
{
    \small
    \bibliographystyle{ieeenat_fullname}
    \bibliography{main}

@String(CVPR= {IEEE Conf. Comput. Vis. Pattern Recog.})

@String(ICCV= {Int. Conf. Comput. Vis.})

@String(ICLR = {Int. Conf. Learn. Represent.})

@String(CVPR  = {CVPR})

@String(ICCV  = {ICCV})

@String(ICLR  = {ICLR})

@article{flashvsr,
  title={FlashVSR: Towards Real-Time Diffusion-Based Streaming Video Super-Resolution},
  author={Zhuang, Junhao and Guo, Shi and Cai, Xin and Li, Xiaohui and Liu, Yihao and Yuan, Chun and Xue, Tianfan},
  journal={arXiv preprint arXiv:2510.12747},
  year={2025}
}

@inproceedings{realvsr,
  title={Real-world video super-resolution: A benchmark dataset and a decomposition based learning scheme},
  author={Yang, Xi and Xiang, Wangmeng and Zeng, Hui and Zhang, Lei},
  booktitle={Proceedings of the IEEE/CVF international conference on computer vision},
  pages={4781--4790},
  year={2021}
}

@inproceedings{realviformer,
  title={Realviformer: Investigating attention for real-world video super-resolution},
  author={Zhang, Yuehan and Yao, Angela},
  booktitle={European conference on computer vision},
  pages={412--428},
  year={2024},
  organization={Springer}
}

@inproceedings{seedvr,
  title={Seedvr: Seeding infinity in diffusion transformer towards generic video restoration},
  author={Wang, Jianyi and Lin, Zhijie and Wei, Meng and Zhao, Yang and Yang, Ceyuan and Loy, Chen Change and Jiang, Lu},
  booktitle={Proceedings of the IEEE/CVF Conference on Computer Vision and Pattern Recognition},
  pages={2161--2172},
  year={2025}
}

@inproceedings{seedvr2,
  title={SeedVR2: One-Step Video Restoration via Diffusion Adversarial Post-Training},
  author ={Wang, Jianyi and Lin, Shanchuan and Lin, Zhijie and Ren, Yuxi and Wei, Meng and Yue, Zongsheng and Zhou, Shangchen and Chen, Hao and Zhao, Yang and Yang, Ceyuan and Xiao, Xuefeng and Loy, Chen Change and Jiang, Lu},
  booktitle={ICLR},
  year={2026}
}

@inproceedings{star,
  title={Star: Spatial-temporal augmentation with text-to-video models for real-world video super-resolution},
  author={Xie, Rui and Liu, Yinhong and Zhou, Penghao and Zhao, Chen and Zhou, Jun and Zhang, Kai and Zhang, Zhenyu and Yang, Jian and Yang, Zhenheng and Tai, Ying},
  booktitle={Proceedings of the IEEE/CVF International Conference on Computer Vision},
  pages={17108--17118},
  year={2025}
}

@article{dove,
  title={Dove: Efficient one-step diffusion model for real-world video super-resolution},
  author={Chen, Zheng and Zou, Zichen and Zhang, Kewei and Su, Xiongfei and Yuan, Xin and Guo, Yong and Zhang, Yulun},
  journal={arXiv preprint arXiv:2505.16239},
  year={2025}
}

@article{rectifiedflow,
  title={Flow straight and fast: Learning to generate and transfer data with rectified flow},
  author={Liu, Xingchao and Gong, Chengyue and Liu, Qiang},
  journal={arXiv preprint arXiv:2209.03003},
  year={2022}
}

@inproceedings{flashattn2,
title={FlashAttention-2: Faster Attention with Better Parallelism and Work Partitioning},
author={Tri Dao},
booktitle={The Twelfth International Conference on Learning Representations},
year={2024},
url={https://openreview.net/forum?id=mZn2Xyh9Ec}
}

@article{selfforcing,
  title={Self Forcing: Bridging the Train-Test Gap in Autoregressive Video Diffusion},
  author={Huang, Xun and Li, Zhengqi and He, Guande and Zhou, Mingyuan and Shechtman, Eli},
  journal={arXiv preprint arXiv:2506.08009},
  year={2025}
}

@inproceedings{causvid,
    title={From Slow Bidirectional to Fast Autoregressive Video Diffusion Models},
    author={Yin, Tianwei and Zhang, Qiang and Zhang, Richard and Freeman, William T and Durand, Fredo and Shechtman, Eli and Huang, Xun},
    booktitle={CVPR},
    year={2025}
}

@article{osediff,
  title={One-step effective diffusion network for real-world image super-resolution},
  author={Wu, Rongyuan and Sun, Lingchen and Ma, Zhiyuan and Zhang, Lei},
  journal={Advances in Neural Information Processing Systems},
  volume={37},
  pages={92529--92553},
  year={2024}
}

@inproceedings{sinsr,
  title={Sinsr: diffusion-based image super-resolution in a single step},
  author={Wang, Yufei and Yang, Wenhan and Chen, Xinyuan and Wang, Yaohui and Guo, Lanqing and Chau, Lap-Pui and Liu, Ziwei and Qiao, Yu and Kot, Alex C and Wen, Bihan},
  booktitle={Proceedings of the IEEE/CVF conference on computer vision and pattern recognition},
  pages={25796--25805},
  year={2024}
}

@inproceedings{seesr,
  title={Seesr: Towards semantics-aware real-world image super-resolution},
  author={Wu, Rongyuan and Yang, Tao and Sun, Lingchen and Zhang, Zhengqiang and Li, Shuai and Zhang, Lei},
  booktitle={Proceedings of the IEEE/CVF conference on computer vision and pattern recognition},
  pages={25456--25467},
  year={2024}
}

@inproceedings{musiq,
  title={Musiq: Multi-scale image quality transformer},
  author={Ke, Junjie and Wang, Qifei and Wang, Yilin and Milanfar, Peyman and Yang, Feng},
  booktitle={Proceedings of the IEEE/CVF international conference on computer vision},
  pages={5148--5157},
  year={2021}
}

@inproceedings{realesrgan,
  title={Real-esrgan: Training real-world blind super-resolution with pure synthetic data},
  author={Wang, Xintao and Xie, Liangbin and Dong, Chao and Shan, Ying},
  booktitle={Proceedings of the IEEE/CVF international conference on computer vision},
  pages={1905--1914},
  year={2021}
}

@inproceedings{lpips,
  title={The unreasonable effectiveness of deep features as a perceptual metric},
  author={Zhang, Richard and Isola, Phillip and Efros, Alexei A and Shechtman, Eli and Wang, Oliver},
  booktitle={Proceedings of the IEEE conference on computer vision and pattern recognition},
  pages={586--595},
  year={2018}
}

@inproceedings{maniqa,
  title={Maniqa: Multi-dimension attention network for no-reference image quality assessment},
  author={Yang, Sidi and Wu, Tianhe and Shi, Shuwei and Lao, Shanshan and Gong, Yuan and Cao, Mingdeng and Wang, Jiahao and Yang, Yujiu},
  booktitle={Proceedings of the IEEE/CVF conference on computer vision and pattern recognition},
  pages={1191--1200},
  year={2022}
}

@article{stablesr,
  title={Exploiting diffusion prior for real-world image super-resolution},
  author={Wang, Jianyi and Yue, Zongsheng and Zhou, Shangchen and Chan, Kelvin CK and Loy, Chen Change},
  journal={International Journal of Computer Vision},
  volume={132},
  number={12},
  pages={5929--5949},
  year={2024},
  publisher={Springer}
}

@inproceedings{diffbir,
  title={Diffbir: Toward blind image restoration with generative diffusion prior},
  author={Lin, Xinqi and He, Jingwen and Chen, Ziyan and Lyu, Zhaoyang and Dai, Bo and Yu, Fanghua and Qiao, Yu and Ouyang, Wanli and Dong, Chao},
  booktitle={European Conference on Computer Vision},
  pages={430--448},
  year={2024},
  organization={Springer}
}

@inproceedings{swinir,
  title={Swinir: Image restoration using swin transformer},
  author={Liang, Jingyun and Cao, Jiezhang and Sun, Guolei and Zhang, Kai and Van Gool, Luc and Timofte, Radu},
  booktitle={Proceedings of the IEEE/CVF international conference on computer vision},
  pages={1833--1844},
  year={2021}
}

@inproceedings{supir,
  title={Scaling up to excellence: Practicing model scaling for photo-realistic image restoration in the wild},
  author={Yu, Fanghua and Gu, Jinjin and Li, Zheyuan and Hu, Jinfan and Kong, Xiangtao and Wang, Xintao and He, Jingwen and Qiao, Yu and Dong, Chao},
  booktitle={Proceedings of the IEEE/CVF conference on computer vision and pattern recognition},
  pages={25669--25680},
  year={2024}
}

@article{wan,
  title={Wan: Open and advanced large-scale video generative models},
  author={Wan, Team and Wang, Ang and Ai, Baole and Wen, Bin and Mao, Chaojie and Xie, Chen-Wei and Chen, Di and Yu, Feiwu and Zhao, Haiming and Yang, Jianxiao and others},
  journal={arXiv preprint arXiv:2503.20314},
  year={2025}
}

@inproceedings{realbasicvsr,
  title={Investigating tradeoffs in real-world video super-resolution},
  author={Chan, Kelvin CK and Zhou, Shangchen and Xu, Xiangyu and Loy, Chen Change},
  booktitle={Proceedings of the IEEE/CVF conference on computer vision and pattern recognition},
  pages={5962--5971},
  year={2022}
}

@article{venhancer,
  title={VEnhancer: Generative Space-Time Enhancement for Video Generation},
  author={He, Jingwen and Xue, Tianfan and Liu, Dongyang and Lin, Xinqi and Gao, Peng and Lin, Dahua and Qiao, Yu and Ouyang, Wanli and Liu, Ziwei},
  journal={arXiv preprint arXiv:2407.07667},
  year={2024}
}

@misc {tae,
  author = {Boer Bohan, Ollin},
  title = {TAEHV: Tiny AutoEncoder for Hunyuan Video},
  year = {2025},
  howpublished = {\url{https://github.com/madebyollin/taehv}},
}

@inproceedings{basicvsr,
  title={Basicvsr: The search for essential components in video super-resolution and beyond},
  author={Chan, Kelvin CK and Wang, Xintao and Yu, Ke and Dong, Chao and Loy, Chen Change},
  booktitle={Proceedings of the IEEE/CVF conference on computer vision and pattern recognition},
  pages={4947--4956},
  year={2021}
}

@inproceedings{basicvsrpp,
  title={Basicvsr++: Improving video super-resolution with enhanced propagation and alignment},
  author={Chan, Kelvin CK and Zhou, Shangchen and Xu, Xiangyu and Loy, Chen Change},
  booktitle={Proceedings of the IEEE/CVF conference on computer vision and pattern recognition},
  pages={5972--5981},
  year={2022}
}

@inproceedings{udm10,
  title={Progressive Fusion Video Super-Resolution Network via Exploiting Non-Local Spatio-Temporal Correlations},
  author={Yi, Peng and Wang, Zhongyuan and Jiang, Kui and Jiang, Junjun and Ma, Jiayi},
  booktitle={IEEE International Conference on Computer Vision (ICCV)},
  pages={3106-3115},
  year={2019}
}

@inproceedings{ultravideo,
  title={UltraVideo: High-Quality UHD Video Dataset with Comprehensive Captions},
  author={Xue, Zhucun and Zhang, Jiangning and Hu, Teng and He, Haoyang and Chen, Yinan and Cai, Yuxuan and Wang, Yabiao and Wang, Chengjie and Liu, Yong and Li, Xiangtai and Tao, Dacheng},
  booktitle={Advances in Neural Information Processing Systems},
  year={2025},
  note={Datasets and Benchmarks Track}
}

@inproceedings{adamw,
  title={Decoupled Weight Decay Regularization},
  author={Loshchilov, Ilya and Hutter, Frank},
  booktitle={International Conference on Learning Representations},
  year={2019}
}

@article{gan,
  title={Generative adversarial nets},
  author={Goodfellow, Ian J and Pouget-Abadie, Jean and Mirza, Mehdi and Xu, Bing and Warde-Farley, David and Ozair, Sherjil and Courville, Aaron and Bengio, Yoshua},
  journal={Advances in neural information processing systems},
  volume={27},
  year={2014}
}

@article{upscaleavideo,
  title={Upscale-{A}-Video: Temporal-consistent diffusion model for real-world video super-resolution},
  author={Zhou, Shangchen and Yang, Peiqing and Wang, Jianyi and Luo, Yihang and Loy, Chen Change},
  journal={IEEE/CVF Conference on Computer Vision and Pattern Recognition},
  pages={2535--2545},
  year={2024}
}

@inproceedings{vgg,
  title={Very Deep Convolutional Networks for Large-Scale Image Recognition},
  author={Simonyan, Karen and Zisserman, Andrew},
  booktitle={International Conference on Learning Representations},
  year={2015}
}

@inproceedings{edvr,
  title={Edvr: Video restoration with enhanced deformable convolutional networks},
  author={Wang, Xintao and Chan, Kelvin CK and Yu, Ke and Dong, Chao and Change Loy, Chen},
  booktitle={Proceedings of the IEEE/CVF conference on computer vision and pattern recognition workshops},
}

@inproceedings{rvrt,
  title={Recurrent Video Restoration Transformer with Guided Deformable Attention},
  author={Liang, Jingyun and Fan, Yuchen and Xiang, Xiaoyu and Ranjan, Rakesh and Ilg, Eddy and Green, Simon and Cao, Jiezhang and Zhang, Kai and Timofte, Radu and Van Gool, Luc},
  booktitle={Advances in Neural Information Processing Systems},
  year={2022}
}

@inproceedings{swin,
  title={Swin transformer: Hierarchical vision transformer using shifted windows},
  author={Liu, Ze and Lin, Yutong and Cao, Yue and Hu, Han and Wei, Yixuan and Zhang, Zheng and Lin, Stephen and Guo, Baining},
  booktitle={Proceedings of the IEEE/CVF international conference on computer vision},
  pages={10012--10022},
  year={2021}
}

@article{vrt,
  title={Vrt: A video restoration transformer},
  author={Liang, Jingyun and Cao, Jiezhang and Fan, Yuchen and Zhang, Kai and Ranjan, Rakesh and Li, Yawei and Timofte, Radu and Van Gool, Luc},
  journal={IEEE Transactions on Image Processing},
  volume={33},
  pages={2171--2182},
  year={2024},
  publisher={IEEE}
}

@article{addsr,
  title={Addsr: Accelerating diffusion-based blind super-resolution with adversarial diffusion distillation},
  author={Tai, Ying and Xie, Rui and Zhao, Chen and Zhang, Kai and Zhang, Zhenyu and Zhou, Jun and Yang, Jian},
  journal={Pattern Recognition},
  pages={113012},
  year={2026},
  publisher={Elsevier}
}

@inproceedings{dmd,
  title={One-step diffusion with distribution matching distillation},
  author={Yin, Tianwei and Gharbi, Micha{\"e}l and Zhang, Richard and Shechtman, Eli and Durand, Fredo and Freeman, William T and Park, Taesung},
  booktitle={Proceedings of the IEEE/CVF conference on computer vision and pattern recognition},
  pages={6613--6623},
  year={2024}
}

@article{dmd2,
  title={Improved distribution matching distillation for fast image synthesis},
  author={Yin, Tianwei and Gharbi, Micha{\"e}l and Park, Taesung and Zhang, Richard and Shechtman, Eli and Durand, Fredo and Freeman, William T},
  journal={Advances in neural information processing systems},
  volume={37},
  pages={47455--47487},
  year={2024}
}

@article{lcm,
  title={Latent consistency models: Synthesizing high-resolution images with few-step inference},
  author={Luo, Simian and Tan, Yiqin and Huang, Longbo and Li, Jian and Zhao, Hang},
  journal={arXiv preprint arXiv:2310.04378},
  year={2023}
}

@inproceedings{zero,
  title={ZeRO: Memory Optimizations Toward Training Trillion Parameter Models},
  author={Rajbhandari, Samyam and Rasley, Jeff and Ruwase, Olatunji and He, Yuxiong},
  booktitle={Proceedings of the International Conference for High Performance Computing, Networking, Storage and Analysis},
  year={2020},
  doi={10.1109/SC41405.2020.00024}
}

@InProceedings{spmc,
  author    = {Xin Tao and
               Hongyun Gao and
               Renjie Liao and
               Jue Wang and
               Jiaya Jia},
  title = {Detail-Revealing Deep Video Super-Resolution},
  booktitle = {The IEEE International Conference on Computer Vision (ICCV)},
  month = {Oct},
  year = {2017}
}

@inproceedings{uformer,
  title={Uformer: A general U-shaped transformer for image restoration},
  author={Wang, Zhendong and Cun, Xiaodong and Bao, Jianmin and Zhou, Wengang and Liu, Jianzhuang and Li, Houqiang},
  booktitle={CVPR},
  pages={17683--17693},
  year={2022}
}

@article{spargeattn2,
  title={SpargeAttention2: Trainable Sparse Attention via Hybrid Top-k+ Top-p Masking and Distillation Fine-Tuning},
  author={Zhang, Jintao and Jiang, Kai and Xiang, Chendong and Feng, Weiqi and Hu, Yuezhou and Xi, Haocheng and Chen, Jianfei and Zhu, Jun},
  journal={arXiv preprint arXiv:2602.13515},
  year={2026}
}

@article{vsa,
  title={Vsa: Faster video diffusion with trainable sparse attention},
  author={Zhang, Peiyuan and Chen, Yongqi and Huang, Haofeng and Lin, Will and Liu, Zhengzhong and Stoica, Ion and Xing, Eric and Zhang, Hao},
  journal={arXiv preprint arXiv:2505.13389},
  year={2025}
}

@inproceedings{deepcache,
  title     = {DeepCache: Accelerating Diffusion Models for Free},
  author    = {Ma, Xinyin and Fang, Gongfan and Wang, Xinchao},
  booktitle = {Proceedings of the IEEE/CVF Conference on Computer Vision and Pattern Recognition (CVPR)},
  pages     = {15762--15772},
  month     = {June},
  year      = {2024}
}

@inproceedings{fastercache,
  title     = {FasterCache: Training-Free Video Diffusion Model Acceleration with High Quality},
  author    = {Lv, Zhengyao and Si, Chenyang and Song, Junhao and Yang, Zhenyu and Qiao, Yu and Liu, Ziwei and Wong, Kwan-Yee K.},
  booktitle = {International Conference on Learning Representations},
  year      = {2025}
}

@inproceedings{toca,
  title     = {Accelerating Diffusion Transformers with Token-wise Feature Caching},
  author    = {Zou, Chang and Liu, Xuyang and Liu, Ting and Huang, Siteng and Zhang, Linfeng},
  booktitle = {International Conference on Learning Representations},
  year      = {2025}
}

@inproceedings{taylorseer,
  title     = {From Reusing to Forecasting: Accelerating Diffusion Models with TaylorSeers},
  author    = {Liu, Jiacheng and Zou, Chang and Lyu, Yuanhuiyi and Chen, Junjie and Zhang, Linfeng},
  booktitle = {Proceedings of the IEEE/CVF International Conference on Computer Vision (ICCV)},
  pages     = {15853--15863},
  year      = {2025}
}

@inproceedings{distrifusion,
  title     = {DistriFusion: Distributed Parallel Inference for High-Resolution Diffusion Models},
  author    = {Li, Muyang and Cai, Tianle and Cao, Jiaxin and Zhang, Qinsheng and Cai, Han and Bai, Junjie and Jia, Yangqing and Li, Kai and Han, Song},
  booktitle = {Proceedings of the IEEE/CVF Conference on Computer Vision and Pattern Recognition (CVPR)},
  pages     = {7183--7193},
  month     = {June},
  year      = {2024}
}

@inproceedings{sageattn,
  title={SageAttention: Accurate 8-Bit Attention for Plug-and-play Inference Acceleration}, 
  author={Zhang, Jintao and Wei, Jia and Zhang, Pengle and Zhu, Jun and Chen, Jianfei},
  booktitle={International Conference on Learning Representations (ICLR)},
  year={2025}
}

@Misc{xformers,
  author =       {Benjamin Lefaudeux and Francisco Massa and Diana Liskovich and Wenhan Xiong and Vittorio Caggiano and Sean Naren and Min Xu and Jieru Hu and Marta Tintore and Susan Zhang and Patrick Labatut and Daniel Haziza and Luca Wehrstedt and Jeremy Reizenstein and Grigory Sizov},
  title =        {xFormers: A modular and hackable Transformer modelling library},
  year =         {2022}
}

@inproceedings{flashattn3,
  title     = {FlashAttention-3: Fast and Accurate Attention with Asynchrony and Low-precision},
  author    = {Shah, Jay and Bikshandi, Ganesh and Zhang, Ying and Thakkar, Vijay and Ramani, Pradeep and Dao, Tri},
  booktitle = {Advances in Neural Information Processing Systems},
  volume    = {37},
  year      = {2024},
  doi       = {10.52202/079017-2193},
}

@article{rollingforcing,
  title={Rolling Forcing: Autoregressive Long Video Diffusion in Real Time},
  author={Liu, Kunhao and Hu, Wenbo and Xu, Jiale and Shan, Ying and Lu, Shijian},
  journal={arXiv preprint arXiv:2509.25161},
  year={2025}
}

@article{liu2025fape,
  title={FAPE-IR: Frequency-Aware Planning and Execution Framework for All-in-One Image Restoration},
  author={Liu, Jingren and Xu, Shuning and Yang, Qirui and Wang, Yun and Chen, Xiangyu and Ji, Zhong},
  journal={arXiv preprint arXiv:2511.14099},
  year={2025}
}
}

\clearpage
\setcounter{page}{1}
\maketitlesupplementary

This supplementary material provides details omitted from the main paper.
\cref{sec:supp_mfsw} describes the MFSWA design, including boundary-clamped gathering and its redundant attention overhead.
\cref{sec:supp_eval} details the unified streaming protocol, additional qualitative results, extended efficiency comparison at \(2560\!\times\!1440\), and cross-backend deployment results.
\cref{sec:supp_limit} summarizes limitations and future directions.

\section{MFSWA Design and Analysis}
\label{sec:supp_mfsw}

The main paper introduces three components of MFSWA: spatial-only partitioning with full temporal visibility, dense-block pre-gathering, and half-window shifting with priority-coherent scattering. This section completes the specification by describing boundary-clamped gathering and its redundant attention cost.

\subsection{Boundary-clamped Gather Overhead}
\label{subsec:supp_gather}
\label{sec:supp_mem}

\paragraph{Construction.}
Window starts are generated by deterministic boundary-clamped indexing. For latent size \(H \times W\) and window size \((w_h, w_w)\), anchors are chosen to cover every spatial location, keep every window at size \(T \cdot w_h \cdot w_w\), and introduce no padding tokens. If \(H\) or \(W\) is not divisible by the window size, boundary indices are clamped so that right or bottom windows overlap adjacent interior windows. The gathered tensor has regular shape \((B \cdot N_w) \times \text{heads} \times (T \cdot w_h \cdot w_w) \times d\), avoiding ragged tensors, padding masks, and variable-size boundary windows.

\paragraph{Why an overhead arises.}
Boundary clamping can place a token in multiple windows, so a layer attends to more than \(H\!\cdot\!W\) spatial tokens. Under the fixed-window implementation, compute is proportional to the total gathered token count. We define \(\alpha\) as the ratio of gathered spatial tokens to \(H\!\cdot\!W\). Thus \(\alpha\) is a compute ratio relative to an ideal equal-size, overlap-free fixed-window partition, not to a ragged boundary implementation whose cost depends on squared boundary-window sizes. Because partitioning is spatial-only, the temporal factor \(T\) cancels. Odd layers dominate the overhead because half-window shifting creates additional boundary overlap.

\paragraph{Coverage factor.}
For one axis of length \(L\) and window size \(w\), even layers use \(n_\text{even} = \lceil L/w \rceil\) windows. Odd layers start with a clamped half-window and cover the remaining \(L-w/2\) locations, giving \(n_\text{odd} = 1 + \lceil (L - w/2)/w \rceil\). The per-axis coverage is \(\rho = n\, w / L\), and the 2D factor is \(\alpha = \rho(H, w_h)\,\rho(W, w_w)\). Applying \(L/w \le \lceil L/w \rceil < L/w + 1\) to each axis,
\[
1 \;\le\; \rho_\text{even} \;<\; 1 + \frac{w}{L},
\qquad
1 + \frac{w}{2L} \;\le\; \rho_\text{odd} \;<\; 1 + \frac{3w}{2L}.
\]
The even-layer factor equals \(1\) exactly when \(L\) is divisible by \(w\). In contrast, \(\rho_\text{odd} > 1\) for all \(L\), because the half-window offset leaves a residual boundary segment. Multiplying across both axes, the odd-layer overhead satisfies
\[
\Big(1 + \tfrac{w_h}{2H}\Big)\Big(1 + \tfrac{w_w}{2W}\Big)
\;\le\; \alpha_\text{odd} \;<\;
\Big(1 + \tfrac{3w_h}{2H}\Big)\Big(1 + \tfrac{3w_w}{2W}\Big).
\]
The bounds depend only on \(w/H\) and \(w/W\). The overhead is content-independent, approaches \(1\) at high resolution, and is largest when a latent axis exceeds a window multiple by about \(w/2\).

\paragraph{Example.}
At \(2560\!\times\!1440\), the latent size is \((H, W) = (45, 80)\) and \((w_h, w_w) = (16, 16)\). Even layers use \(3 \times 5\) windows, giving \(\alpha_\text{even} = \tfrac{48}{45}\cdot\tfrac{80}{80} \approx 1.07\). Odd layers use \(4 \times 6\) windows, giving \(\alpha_\text{odd} = \tfrac{64}{45}\cdot\tfrac{96}{80} \approx 1.71\). At \(3840\!\times\!2160\), the latent size is \((68, 120)\), and the odd layer uses \(5 \times 8\) windows, giving \(\alpha_\text{odd}=\tfrac{80}{68}\cdot\tfrac{128}{120}\approx1.255\).

\paragraph{Relation to measured memory.}
The coverage factor \(\alpha\) describes redundant attention compute, not peak memory. Since the gathered \(Q,K,V\) windows are transient SDPA inputs rather than persistent activations, the odd-layer overhead of \(\alpha_\text{odd}\approx1.71\) at \(2560\!\times\!1440\) does not imply a comparable memory increase. In practice, peak memory is dominated by resident activations and workspace, yielding only a modest \(35.37\!\to\!38.01\)~GB increase in Table~\ref{tab:ablation_mfsw}. The extra attention remains bounded by the resolution and window size, and decreases at higher resolutions.

\subsection{Dense SDPA Implementation}
\label{subsec:supp_props}
With these components, each window uses one dense SDPA call. The window layout is encoded by two precomputed index tensors cached per resolution. The training graph contains no attention mask, padding token, block-sparse descriptor, or cyclic shift. MFSWA obtains locality from the partition while keeping all attention calls dense.

\section{Evaluation and Deployment}
\label{sec:supp_eval}

This section specifies the unified streaming protocol, additional qualitative results, extended efficiency comparison at \(2560\!\times\!1440\), and the cross-backend deployment results.

\subsection{Unified Streaming Evaluation Protocol}
\label{subsec:supp_protocol}
\label{sec:supp_protocol}

Table~\ref{tab:main} requires a like-for-like streaming evaluation. Because the baselines use different temporal strides and overlap conventions, we use a unified protocol. RealBasicVSR and RealViFormer process \(24\)-frame chunks with a \(4\)-frame overlap, and metrics are computed only on non-overlapped outputs. Upscale-A-Video, SeedVR2-3B, and DOVE process \(25\)-frame chunks (\(=\!4k+1\)) with a \(4\)-frame overlap. Real-ESRGAN and FlashVSR-Tiny use their official evaluation scripts, as they already operate per frame or per causal block. SwiftVR uses its native causal chunk protocol without overlap, and ReAE carries boundary states across chunks. All methods use the same input resolution and test clips. Metrics are computed only on emitted frames. We use official implementations and released default precision: \texttt{float32} for non-diffusion baselines and \texttt{bfloat16} for diffusion-based methods.

This protocol supports both the quality results in Table~\ref{tab:main} and the efficiency results in Table~\ref{tab:portability}. It evaluates all methods under the same streaming constraint, so the numbers may differ from the original offline reports. Chunking also improves efficiency because attention cost scales quadratically with temporal length.

\subsection{Additional Visualization Results}
\label{subsec:supp_visual}

\begin{figure*}[t]
    \centering
    \includegraphics[width=\linewidth]{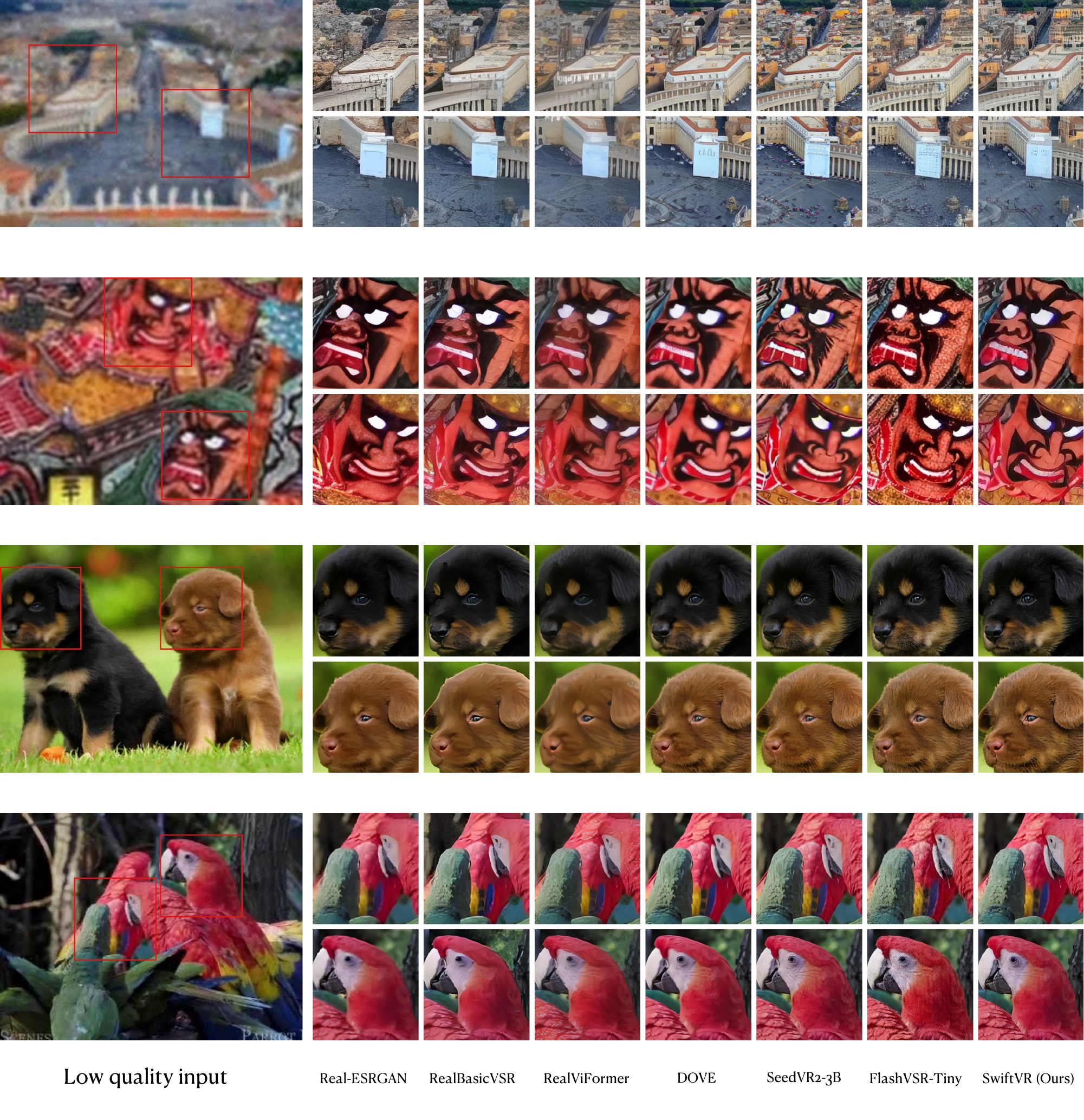}
    \caption{Additional qualitative comparisons on real world videos. Columns show the low quality input, Real-ESRGAN, RealBasicVSR, RealViFormer, DOVE, SeedVR2-3B, FlashVSR-Tiny, and SwiftVR (Ours).}
    \label{fig:sup_fig}
\end{figure*}

Figure~\ref{fig:sup_fig} presents additional qualitative comparisons on real world videos. The examples include distant buildings, mural patterns, animal fur, and bird plumage. Regression based methods recover coarse structures but smooth fine details and reduce local contrast. DOVE produces stable outputs but preserves less high frequency detail. SeedVR2-3B and FlashVSR-Tiny recover sharper patterns, but may introduce color shifts, halos, or excessive sharpening. SwiftVR restores clearer boundaries and more natural details, including roof edges, fur, and feather structures, with stable color and fewer local artifacts. These results further support the perceptual gains shown in the main comparison.

\subsection{Extended Per-method Efficiency Comparison}
\label{subsec:supp_eff}
\label{sec:supp_eff}

Table~\ref{tab:portability_full} extends the \(2560\!\times\!1440\) efficiency comparison by adding non-generative baselines to the one-step diffusion methods. Upscale-A-Video is excluded from this timing table because it is a 30-step baseline, but it remains included in quality evaluation and the 4K OOM check.

\begin{table}[htbp]
\centering
\small
\caption{Extended efficiency comparison at \(2560\!\times\!1440\) on one H100 under causal streaming, measured over \(24\) output frames. The table includes non-generative baselines and one-step diffusion methods; SeedVR2-3B and DOVE use \texttt{use\_tile=True}.}
\label{tab:portability_full}
\setlength{\tabcolsep}{3pt}
\resizebox{\linewidth}{!}{
\begin{tabular}{l|ccc|cccc}
\toprule
Metric & Real-ESRGAN & RealBasicVSR & RealViFormer & SeedVR2-3B & DOVE & FlashVSR-Tiny & SwiftVR (Ours) \\
\midrule
Params (M) & 16.70 & 6.29 & 5.82 & 3{,}642 & 10{,}548 & 1{,}752 & 5{,}081 \\
Avg. Time (s) & 5.770 & 1.324 & 0.940 & 17.320 & 27.615 & 2.493 & 0.766 \\
FPS & 4.16 & 18.12 & 25.53 & 1.39 & 0.87 & 9.61 & 31.32 \\
Peak Mem. (GB) & 3.15 & 7.12 & 5.82 & 35.35 & 59.24 & 34.35 & 38.01 \\
\bottomrule
\end{tabular}}
\end{table}

At \(3840\!\times\!2160\), all compared one-step diffusion-based VR methods run out of memory on a single H100-80G under the same streaming protocol, even with VAE tiling enabled. SwiftVR sustains \(13.84\)~FPS at this resolution with peak memory of \(60.91\)~GB (Table~\ref{tab:runtime_breakdown}). The non-generative baselines (Real-ESRGAN, RealBasicVSR, RealViFormer) do fit at \(3840\!\times\!2160\) but operate at substantially lower perceptual quality, as already shown in Table~\ref{tab:main} at the standard test resolutions.

\subsection{Cross-backend Deployment}
\label{subsec:supp_backend}
\label{sec:supp_backend}

MFSWA keeps every attention call on the standard dense SDPA interface, so SwiftVR can run on different fused-attention backends without weight conversion. Table~\ref{tab:ablation_backend} reports throughput for five backends at \(2560\!\times\!1440\). Peak memory remains \(38.01\)~GB and metrics match to the reported precision, so both are omitted.

\begin{table}[!h]
\small
\centering
\caption{Cross-backend deployment on one H100 at \(2560\!\times\!1440\). Peak memory is constant at \(38.01\)~GB and restoration metrics match to the reported precision.}
\label{tab:ablation_backend}
\setlength{\tabcolsep}{8pt}
\renewcommand{\arraystretch}{1.1}
\begin{tabular}{lc}
\toprule
Attention backend & FPS$\uparrow$ \\
\midrule
PyTorch SDPA (cuDNN/Flash auto) & 31.32 \\
FlashAttention-2~\cite{flashattn2} & 31.36 \\
FlashAttention-3~\cite{flashattn3} & 32.31 \\
SageAttention~\cite{sageattn}      & 30.70 \\
xFormers~\cite{xformers}           & 31.36 \\
\bottomrule
\end{tabular}
\end{table}

On H100, PyTorch SDPA already selects the cuDNN/Flash path and matches FlashAttention-2 and xFormers within about \(0.1\%\). FlashAttention-3 is about \(3\%\) faster than SDPA. SageAttention is slightly slower at this scale and precision, although it can outperform FlashAttention on Ada-class consumer GPUs~\cite{sageattn}. These results mainly confirm backend portability. MFSWA preserves dense SDPA compatibility while introducing window locality.

\section{Limitations and Future Work}
\label{sec:supp_limit}
\label{sec:supp_limitations}

\paragraph{Limitations.}
SwiftVR does not yet deliver real-time generative 4K restoration on consumer GPUs. At \(3840\!\times\!2160\), it reaches \(13.84\)~FPS with \(60.91\)~GB peak memory on an H100. This fits a server GPU but exceeds consumer-GPU memory and remains below \(24\)~FPS. Real-time 4K restoration on consumer GPUs remains future work.

\paragraph{Future work.}
SwiftVR currently uses no inference-side acceleration. Future work will target two directions. The first is inference acceleration, including post-training quantization, KV-state caching and compression, and learned token reduction, all of which are orthogonal to the architecture. The second is a smaller, more compressed backbone. Wan2.2-TI2V-5B remains large, so higher latent compression and smaller base models are likely necessary for real-time 4K on consumer GPUs.

\end{document}